# Reasoning in a Hierarchical System with Missing Group Size Information

Subhash Kak

**Abstract.**
Reasoning under uncertainty is basic to any artificial intelligence system designed to emulate human decision making. This paper analyzes the problem of judgments or preferences subsequent to initial analysis by autonomous agents in a hierarchical system. We first examine a method where comparisons between two alternatives are made across different groups, and we show that this method reduces instances of preference reversal of the kind encountered in Simpson's paradox. It is also shown that comparing across all permutations gives better results than exhaustive pairwise comparison. We next propose a method where the preference weights are recomputed based on relative group size values and show that it works well. This problem is also of interest for preventing fraud where the data that was obtained all together is presented in groups in a manner that supports a hypothesis that is reverse of the one valid for the entire group. We also discuss judgments where more than two choices are involved.

Keywords: hierarchical reasoning systems, group choice, probability reversal, decision theory

**Introduction**
In artificial intelligence systems, decisions must be made after initial processing of the raw data that reduces its dimensions through appropriate methods of representation [1][2]. The look-ahead human reasoning is accomplished under conditions of uncertainty and that needs to be emulated by artificial agents [3], where the logic is so designed that it mitigates the effects of the uncertainty.

Here we consider the situation where the uncertainty is specifically concerning the sizes of the groups and the decision is to choose between alternatives. Such decision making is also a characteristic of responses to surveys and voting systems where the respondents' decision is made based on prior bias or incomplete information. Probability computations based on frequency of an outcome on a small group can change when data from more than one cohort are aggregated, as is true of the Simpson's paradox [4][5].

This problem is significant in human decision making and responses to surveys because of uncertainty of knowledge and aspects related to attention [6][7]. Cognitive processing may also lead to preference reversal [8][9]. Human judgments may be modeled to have a superpositional component that is quite like a quantum system [10][11][12] that provides insight into human decision making as in the disjunction effect. It should be noted that this comes with associated limitations [13][14].

This paper considers agents in a hierarchical system where the higher level must choose between alternatives where it does not possess all the data that went in the alternatives picked by the lower level agents. It is shown how a strategy of mixing inputs from different agents and having the lower agents work in the size discrepancies in their preferences can improve the decision by the higher agent. A method to recompute the preferences is also presented and it is shown that it improves the higher level agent results.

**The Hierarchical Decision System**

Consider the hierarchical information system of Figure 1, in which the decision by D is made on partially processed data by agents $A_1$ through $A_n$.

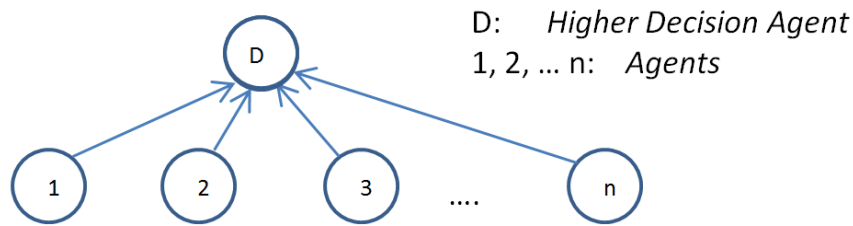

D: *Higher Decision Agent*
1, 2, … n: *Agents*

Figure 1. A hierarchical decision-system

Let the raw data arriving at the agents $A_1$ through $A_n$ be called $M_i$. Agent *i* performs $O_i$ so that it obtains $U_i$:

$$O_i(M_i) = U_i$$

and, doing so, reduces the dimensions and size of the data. The higher level decision agent, D, operates on the $U_i$ and obtains $J_k$, k=1, 2, for a binary preference system:

$$J_k = S(U_1, U_2, …, U_n), k=1, 2$$

To provide an example, assume that there are 3 agents that receive information on performance in trials related to two treatments, that we call Treatment 1 and Treatment 2, respectively. Table 1 gives the data on the results:

Table 1. An experiment with three agents

|  | *Agent 1* | *Agent 2* | *Agent 3* | Totals |
|---|---|---|---|---|
| Treatment 1 | 0.00 (0/1) | 0.75 (3/4) | 0.60 (3/5) | 0.60 (6/10) |
| Treatment 2 | 0.20 (1/5) | 1.00 (1/1) | 0.75 (3/4) | 0.50 (5/10) |

Each agent judges Treatment 2 to be superior. The frequency differences are substantial and equal to 0.20, 0.25, and 0.15, respectively. However, when the preferences of the three agents are aggregated by the higher decision agent, Treatment 1 scores over Treatment 2 by 0.10 points.

The preference reversal is due to the that if $\frac{a_1}{b_1} > \frac{c_1}{d_1}, \frac{a_2}{b_2} > \frac{c_2}{d_2}$ and $\frac{a_3}{b_3} > \frac{c_3}{d_3}$ there will be "pathological" situations so that and $\frac{c_1+c_2+c_3}{d_1+d_2+d_3} > \frac{a_1+a_2+a_3}{b_1+b_2+b_3}$. In general, there would be such "pathological" cases irrespective of the number of such inequalities.

The numbers involved for each of the agents is kept small just for convenience and there could be larger numbers of the kind encountered in testing for efficacy of one treatment over another, or comparison with a placebo [15]. Typically, the cohorts in such testing do not have identical size because many volunteers drop out before the treatment has run its course or due to the difficulty of getting the volunteers to enlist.

For a larger example, see Table 2, where the data is in terms of six sets of groups (dealt with by different agents). For each of the agents Treatment 2 is superior to Treatment 1 by margins that range from 0.04 to 0.13, but when the data is aggregated by the higher level decision agent, Treatment 1 scores over Treatment 2 by a margin of 0.06.

Table 2. An experiment with six agents

|  | Agent 1 | Agent 2 | Agent 3 | Agent 4 | Agent 5 | Agent 6 | Totals |
|---|---|---|---|---|---|---|---|
| Treatment 1 | 0.62 (5/8) | 0.85 (23/27) | 0.60 (15/24) | 0.84 (68/81) | 0.69 (55/80) | 0.87 (234/270) | 0.82 (400/490) |
| Treatment 2 | 0.73 (19/26) | 0.89 (8/9) | 0.73 (57/78) | 0.85 (23/27) | 0.73 (192/263) | 0.93 (81/87) | 0.76 (380/490) |

The frequency of occurrence of Simpson's paradox is known for different assumptions made about the data. In in a random 2 × 2 × 2 table with uniform distribution, the paradox occurs with a probability of 1/60 [16], whereas in random path models under certain conditions it is a bit higher than 1/8 [17]. Further issues about preference reversal arise that are discussed in [18],[19].

In general, if the decision agent D has data in categories provided as in Table 3. Let $s_i$ be the outcomes out of a total of $e_i$ for Treatment 1 and $p_i$ be the corresponding outcomes out of a total of $f_i$ for Treatment 2, where the outcome is to be interpreted as the proportion of subjects who improved within the cohort.

Table 3. Agent preferences represented directly

|  | Agent 1 | Agent 2 | ...... | Agent n |
|---|---|---|---|---|
| Treatment 1 | $s_1/g_1$ | $s_2/g_2$ | ...... | $s_n/g_n$ |
| Treatment 2 | $p_1/h_1$ | $p_2/h_2$ | ...... | $p_n/h_n$ |

It is natural to have the following decision system:

$$\sum_i s_i/g_i > \sum_i p_i/h_i, \text{ then Treatment 1} > \text{Treatment 2}$$

When n>2, the decision may also be made by majority logic.

These ideas are also relevant to voting schemes {20][21][22], where $s_i/g_i$ and $p_i/h_i$ represent preference values for the two different outcomes.

**Comparisons across Sets**

We first consider comparison in pairs, allowing switching of the responses. We see that between agents 1 and 2 the data becomes

| 0.00 (0/1) | 0.75 (3/4) |
|---|---|
| 0.20 (1/5) | 1.00 (1/1) |

which when switched yields a result of 1 -1. Likewise, for 2 and 3, it yields

| 0.75 (3/4) | 0.60 (3/5) |
|---|---|
| 1.00 (1/1) | 0.75 (3/4) |

This is one tie and 2 better in one case. For 1 and 3, we have

| 0.00 (0/1) | 0.60 (3/5) |
|---|---|
| 0.20 (1/5) | 0.75 (3/4) |

This again yields a tie.

So switching data in pairs shows that preferences for Treatment 2 and Treatment 1 are not that different in their efficacy.

In comparison across sets, the decision agent D at the higher level uses not just the preference values that have been provided to it for that specific group. Therefore, if the outputs of the Agents 1, 2, and 3 are called $A_1$, $B_1$, $C_1$, respectively in favor of Treatment 1, and $A_2$, $B_2$, $C_2$, respectively, in favor of Treatment 2, then they should be compared with each other in all possible arrangements, of which there will be 6 in this example. In other words $A_1$, $B_1$, $C_1$ should be compared not only with $A_2$, $B_2$, $C_2$ but also with

$A_2$, $C_2$, $B_2$
$B_2$, $C_2$, $A_2$
$B_2$, $A_2$, $C_2$
$C_2$, $A_2$, $B_2$
$C_2$, $B_2$, $A_2$

The full table for our example will be as follows in Table 4, where the first row represents frequency values for Treatment 1 for the three cohorts and the remaining rows represent the frequency values for Treatment 2 in different permutations with respect to Treatment 1.

Table 4. Comparison across groups

| 0 | 0.36 | 0.75 |
|---|------|------|
| **0.20** | 0.60 | **1.00** |
| **0.20** | 1.00 | 0.60 |
| **1.00** | 0.20 | 0.60 |
| **1.00** | 0.60 | 0.20 |
| **0.60** | 1.00 | 0.20 |
| **0.60** | 0.20 | **1.00** |

We see that when the comparisons are made in all arrangements, Treatment 2 scores 12 out of 18 times (the recomputed values are shown in bold). Therefore, its favorability is not as absolute as when shown in Table 1.

**More than two alternatives**

We now consider multiple alternatives. Consider 3 alternatives as below in Table 5 where the total number of subjects is not the same in the two groups; however the number of subjects for testing each alternative is the same.

Table 5. Example with three alternatives

|   | Group 1 | Group 2 | Totals |
|---|---------|---------|--------|
| A | 1/10 (0.10) | 69/90 (0.77) | 70/100 (0.70) |
| B | 10/50 (0.20) | 40/50 (0.80) | 50/100 (0.50) |
| C | 22/80 (0.27) | 18/20 (0.90) | 40/100 (0.40) |

Even though the best performance for data aggregated across the groups is for A, followed by B and C (see third column of Table 5), the results are reversed within each group, so that for each of them separately, C scores the highest, followed by B and A, in that order.

**Accounting for difference in group samples**

We seek a function that leaves the preferences unchanged if the sample size is equal to the mean or is larger, and penalizes preferences obtained from samples that are smaller than the mean, with the penalty increasing as the departure from the mean becomes larger. Many different functions may be used; here we propose one that corresponds to Figure 1.

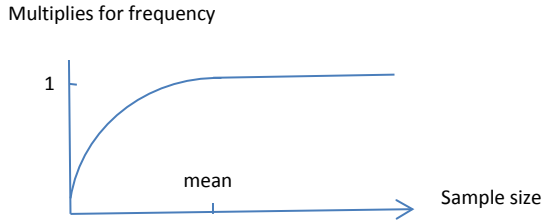

Figure 1. Recomputation function shape

If the preference weight s is obtained from the group of size g (where the mean of all the groups is $g_m$), then its value should be adjusted for the differences in group sizes in this general form:

$$s_{adjusted} = \begin{cases} 1, & \text{when } g \geq g_m \\ kse^{-W(g-g_m)}, & \text{when } g < g_m \end{cases} \quad (1)$$

where $W(g-g_m)$ is some appropriate function and where k is some suitably chosen constant. An attractive choice of this function is in equation (2) for it is quadratic and it gives us the value of 1 at the edge of $g = g_m$:

$$W(g - g_m) = \frac{\delta(g - g_m)^2}{\text{Variance of } g} \quad (2)$$

But here, for convenience, we use a δ= ½, k=1, so that equation (1) transforms into

$$s_{adjusted} = \begin{cases} 1, & \text{when } g \geq g_m \\ se^{-(g-g_m)^2/2g_{variance}}, & \text{when } g < g_m \end{cases} \quad (3)$$

Thus we see that the recomputation weight is $e^{-0.5v^2}$ if the $g$ is $v$ standard deviations away from the mean. So if the test size is one standard deviation away, the preference value should be reduced by 0.60; if it is 2 standard deviations away, it should be reduced by 0.14; for 3 standard deviations away, by 0.011, and so on.

In general, the choice of the function may be based on the nature of the data so that the comparison between the preferences is intuitively satisfactory. But the problem of alternative recomputation functions shall not be considered here.

Given equation (1), we now recompute the preference frequencies of Tables 1, 2 and 4. In Table 1, the values of Agent 1 require no change since the first is 0 and the second is for a group whose size exceeds the mean of 3. For Agent 2, the first row is fine as 4 is larger than the mean of 2.5. The variance is 8.5-6.25 = 2.25 and so the second entry will become $\exp(-1.5^2/2 \times 2.25) = 0.60$. The second entry for Agent 3, likewise becomes $0.75 \times \exp(-0.5^2/2 \times 0.25) = 0.45$ (since the variance for 4 and 5 is 0.25).

Table 6. Recomputed preference frequencies using Equation (3)

|  | Agent 1 | Agent 2 | Agent 3 |
|---|---|---|---|
| Treatment 1 | 0.00 (0/1) | 0.75 (3/4) | 0.60 (3/5) |
| Treatment 2 | 0.20 (1/5) | **0.36** (1/1) | **0.45** (3/4) |

The recomputed values are shown in bold. We find that for Agent 1, Treatment 2 remains superior; Treatment 1 is much better for Agents 2 and 3. This is satisfactory analysis and quite in agreement with our intuition.

Table 7. Recomputing of Table 2 preferences (recomputed values are bold)

|  | Agent 1 | Agent 2 | Agent 3 | Agent 4 | Agent 5 | Agent 6 |
|---|---|---|---|---|---|---|
| Treatment 1 | **0.38** (5/8) | 0.85 (23/27) | **0.36** (15/24) | 0.84 (68/81) | **0.42** (55/80) | 0.87 (234/270) |
| Treatment 2 | 0.73 (19/26) | **0.54** (8/9) | 0.73 (57/78) | **0.52** (23/27) | 0.73 (192/263) | **0.56** (81/87) |

Unlike for the original data where Treatment 2 scored over Treatment 1 for all the agents, now Treatment 1 and Treatment 2 are tied 3-3, which is much more reasonable and not likely to be misinterpreted.

Now let us consider the example with three alternatives of Table 5. The data of Group 1 has a mean of 46.66, therefore the recomputation is required to be done only for the entry for A, whereas the data for Group 2 has a mean of 53.33, so we must recompute values for B and C; note the variances are 2177 and 8156, respectively. With this information, we can use Equation 3 to revise the figures and these are in Table 8.

Table 8. Recomputing Table 5 preference values

|   | Group 1 | Group 2 |
|---|---------|---------|
| A | 1/10 (**0.098**) | 69/90 (0.77) |
| B | 10/50 (0.20) | 40/50 (**0.68**) |
| C | 22/80 (0.27) | 18/20 (**0.88**) |

Once again, it provides preferences results that are superior to those where the raw data is used, as done earlier.

Lastly, we take a new example of choice between 3 alternatives, where the raw figures are 2/4, 3/7 and 4/10, with corresponding preference values of 0.5, 0.43, and 0.40. The mean of the group sizes is 7 and the variance is 6. We can now obtain the corrections to the figures for group that is smaller than the mean, which is the first one. After correction, its value drops to 0.416 and now we find it is not the best of the three.

**Conclusions**

This paper analyzes the problem of arriving at probability judgments subsequent to initial analysis by autonomous agents in a hierarchical system. From the perspective of the higher agent, this represents the problem of decision making under a degree of uncertainty, which is typical in big data with lower-level processing farmed out to different contractors. We propose a method where the comparisons are made across different cohorts, and we show that this reduces instances of preference reversal of the kind encountered in Simpson's paradox. It is also shown that comparing across all permutations gives better results than exhaustive pairwise comparison. We also propose a method where the preference weights are recomputed based on relative group size values.

The problem of preference reversal is also of interest for preventing fraud where the data that was obtained all together is manipulated [20] and presented in groups in a manner that supports a hypothesis that is reverse of the one valid for the entire group. It is also of interest in economic systems where one is interested in why agents behave in a particular fashion [21][22]. It is also of interest in voting systems where such voting is cast independently in different groups, as would be the case in surveys.